\documentclass[10pt,twocolumn,letterpaper]{article}

\usepackage{cvpr}
\usepackage{times}
\usepackage{epsfig}
\usepackage{graphicx}
\usepackage{amsmath}
\usepackage{amssymb}
\usepackage{comment}
\usepackage{caption}
\usepackage{stfloats}
\usepackage{url}

\usepackage{booktabs}
\usepackage{array}
\newcolumntype{P}[1]{>{\centering\arraybackslash}p{#1}}
\newcolumntype{M}[1]{>{\centering\arraybackslash}m{#1}}
\usepackage[pagebackref=true,breaklinks=true,letterpaper=true,colorlinks,bookmarks=false]{hyperref}

\cvprfinalcopy 

\def\cvprPaperID{xxxx} 

\begin{document}

\newcommand{\OURS}{3D-SIS}

\title{\OURS: 3D Semantic Instance Segmentation of RGB-D Scans}

\author{
\parbox{4cm}{\centering Ji Hou\quad}
\parbox{4cm}{\centering Angela Dai\quad}
\parbox{4cm}{\centering Matthias Nie{\ss}ner}\\[0.3em]
Technical University of Munich
\vspace{0.2cm}
}

\twocolumn[{%
\renewcommand\twocolumn[1][]{#1}%
\maketitle

\begin{center}
\vspace{-0.6cm}
\includegraphics[width=\linewidth]{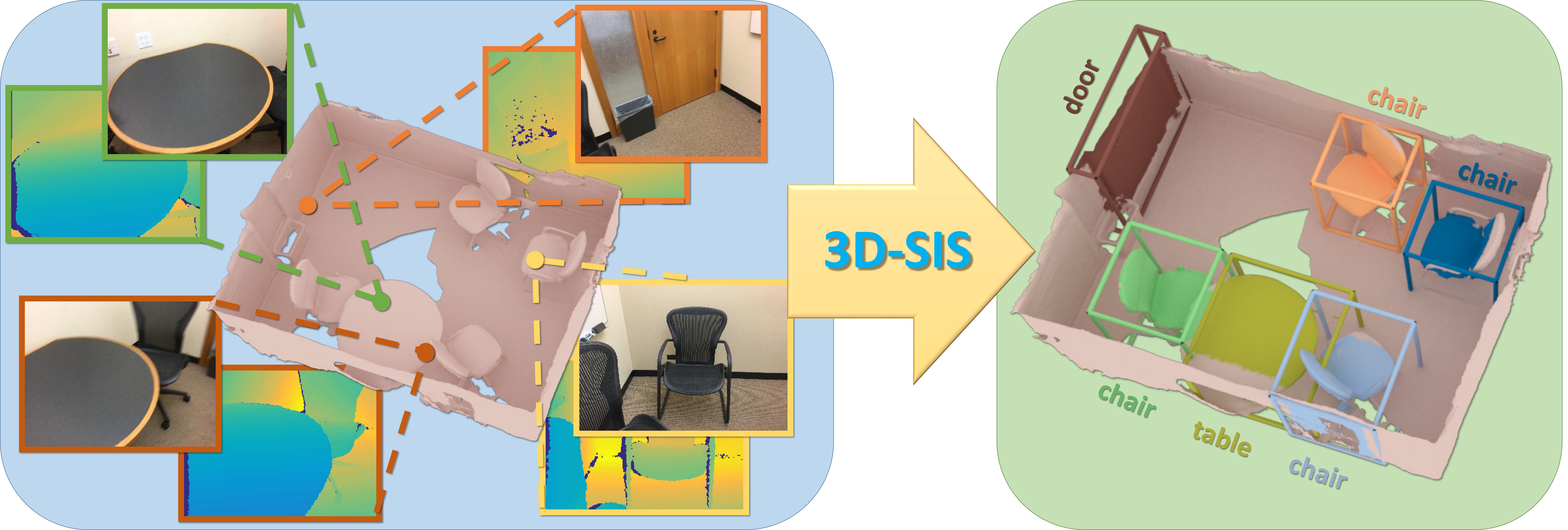}
\captionof{figure}{\OURS{} performs 3D instance segmentation on RGB-D scan data, learning to jointly fuse both 2D RGB input features with 3D scan geometry features. In combination with a fully-convolutional approach enabling inference on full 3D scans at test time, we achieve accurate inference for object bounding boxes, class labels, and instance masks.
}
\label{fig:teaser}
\vspace{0.11cm}
\end{center}
}]
\begin{abstract}
We introduce \OURS{}\footnote{Code available: \url{https://github.com/Sekunde/3D-SIS}}, a novel neural network architecture for 3D semantic instance segmentation in commodity RGB-D scans. The core idea of our method is to jointly learn from both geometric and color signal, thus enabling accurate instance predictions. Rather than operate solely on 2D frames, we observe that most computer vision applications have multi-view RGB-D input available, which we leverage to construct an approach for 3D instance segmentation that effectively fuses together these multi-modal inputs. Our network leverages high-resolution RGB input by associating 2D images with the volumetric grid based on the pose alignment of the 3D reconstruction. For each image, we first extract 2D features for each pixel with a series of 2D convolutions; we then backproject the resulting feature vector to the associated voxel in the 3D grid. This combination of 2D and 3D feature learning allows significantly higher accuracy object detection and instance segmentation than state-of-the-art alternatives. We show results on both synthetic and real-world public benchmarks, achieving an improvement in mAP of over 13 on real-world data.
\end{abstract}

\section{Introduction}

Semantic scene understanding is critical to many real-world computer vision applications.
It is fundamental towards enabling interactivity, which is core to robotics in both indoor and outdoor settings, such as autonomous cars, drones, and assistive robotics, as well as upcoming scenarios using mobile and AR/VR devices.
In all these applications, we would not only want semantic inference of single images, but importantly, also require understanding of spatial relationships and layouts of objects in 3D environments.

With recent breakthroughs in deep learning and the increasing prominence of convolutional neural networks, the computer vision community has made tremendous progress on analyzing images in the recent years.
Specifically, we are seeing rapid progress in the tasks of semantic segmentation \cite{long2015fully,iandola2014densenet,paszke2016enet}, object detection \cite{girshick2015fast,redmon2016you}, and semantic instance segmentation \cite{he2017mask}.
The primary focus of these impressive works lies in the analysis of visual input from a single image;
however, in many real-world computer vision scenarios, we rarely find ourselves in such a single-image setting.
Instead, we typically record video streams of RGB input sequences, or as in many robotics and AR/VR applications, we have 3D sensors such as LIDAR or RGB-D cameras.

In particular, in the context of semantic instance segmentation, it is quite disadvantageous to run methods independently on single images given that instance associations must be found across a sequence of RGB input frames.
Instead, we aim to infer spatial relationships of objects as part of a semantic 3D map, learning prediction of spatially-consistent semantic labels and the underlying 3D layouts {\em jointly} from all input views and sensor data.
This goal can also be seen as similar to traditional sensor fusion but for deep learning from multiple inputs.

We believe that robustly-aligned and tracked RGB frames, and even depth data, from SLAM and visual odometry provide a unique opportunity in this regard.
Here, we can leverage the given mapping between input frames, and thus learn features jointly from all input modalities.
In this work, we specifically focus on predicting 3D semantic instances in RGB-D scans, where we capture a series of RGB-D input frames (e.g., from a Kinect Sensor), compute 6DoF rigid poses, and reconstruct 3D models.
The core of our method learns semantic features in the 3D domain from both color features, projected into 3D, and geometry features from the signed distance field of the 3D scan. 
This is realized by a series of 3D convolutions and ResNet blocks.
From these semantic features, we obtain anchor bounding box proposals.
We process these proposals with a new 3D region proposal network (3D-RPN) and 3D region of interest pooling layer (3D-RoI) to infer object bounding box locations, class labels, and per-voxel instance masks.
In order to jointly learn from RGB frames, we leverage their pose alignments with respect to the volumetric grid.
We first run a series of 2D convolutions, and then backproject the resulting features into the 3D grid.
In 3D, we then join the 2D and 3D features in end-to-end training constrained by bounding box regression, object classification, and semantic instance mask losses.

Our architecture is fully-convolutional, enabling us to efficiently infer predictions on large 3D environments in a single shot.
In comparison to state-of-the-art approaches that operate on individual RGB images, such as Mask R-CNN \cite{he2017mask}, our approach achieves significantly higher accuracy due to the joint feature learning.

To sum up, our contributions are the following:
\begin{itemize} \itemsep0em 
    \item We present the first approach leveraging joint 2D-3D end-to-end feature learning on both geometry and RGB input for 3D object bounding box detection and semantic instance segmentation on 3D scans.
    \item We leverage a fully-convolutional 3D architecture for instance segmentation trained on scene parts, but with single-shot inference on large 3D environments.
    \item We outperform state-of-the-art by a significant margin, increasing the mAP by 13.5 on real-world data.
\end{itemize}
\section{Related Work}

\subsection{Object Detection and Instance Segmentation}
With the success of convolutional neural network architectures, we have now seen impressive progress on object detection and semantic instance segmentation in 2D images~\cite{girshick2015fast,ren2015faster,liu2016ssd,redmon2016you,lin2017feature,he2017mask,lin2018focal}.
Notably, Ren et al.~\cite{ren2015faster} introduced an anchor mechanism to predict objectness in a region and regress associated 2D bounding boxes while jointly classifying the object type.
Mask R-CNN~\cite{he2017mask} expanded this work to semantic instance segmentation by predicting a per-pixel object instance masks.
An alternative direction for detection is the popular Yolo work~\cite{redmon2016you}, which also defines anchors on grid cells of an image. 

This progress in 2D object detection and instance segmentation has inspired work on object detection and segmentation in the 3D domain, as we see more and more video and RGB-D data become available.
Song et al. proposed Sliding Shapes to predict 3D object bounding boxes from single RGB-D frame input with handcrafted feature design~\cite{song2014sliding}, and then expanded the approach to operate on learned features~\cite{song2015deep}.
The latter direction leverages the RGB frame input to improve classification accuracy of detected objects; in contrast to our approach, there is no explicit spatial mapping between RGB and geometry for joint feature learning.
An alternative approach is taken by Frustum PointNet~\cite{qi2017frustum}, where detection is performed a 2D frame and then back-projected into 3D from which final bounding box predictions are refined. 
Wang et al.~\cite{wang2018sgpn} base their SGPN approach on semantic segmentation from a PointNet++ variation.
They formulate instance segmentation as a clustering problem upon a semantically segmented point cloud by introducing a similarity matrix prediction similar to the idea behind panoptic segmentation \cite{kirillov2018panoptic}.
In contrast to these approaches, we explicitly map both multi-view RGB input with 3D geometry in order to jointly infer 3D instance segmentation in an end-to-end fashion.

\subsection{3D Deep Learning}
In the recent years, we have seen impressive progress in developments on 3D deep learning.
Analogous to the 2D domain, one can define convolution operators on volumetric grids, which for instance embed a surface representation as an implicit signed distance field~\cite{curless1996volumetric}.
With the availability of 3D shape databases \cite{wu20153d,chang2015shapenet,song2017ssc} and annotated RGB-D datasets \cite{song2015sun,armeni2017joint,dai2017scannet,chang2017matterport3d}, these network architectures are now being used for 3D object classification \cite{wu20153d,maturana2015voxnet,qi2016volumetric,riegler2017octnet}, semantic segmentation \cite{dai2017scannet,tatarchenko2018tangent,dai20183dmv}, and object or scene completion \cite{dai2017shape,song2017ssc,dai2018scancomplete}.
An alternative representation to volumetric grids are the popular point-based architectures, such as PointNet~\cite{qi2017pointnet} or PointNet++~\cite{qi2017pointnet++}, which leverage a more efficient, although less structured, representation of 3D surfaces.
Multi-view approaches have also been proposed to leverage RGB or RGB-D video information. Su et al. proposed one of the first multi-view architectures for object classification by view-pooling over 2D predictions~\cite{su2015multi}, and Kalogerakis et al. recently proposed an approach for shape segmentation by projecting predicted 2D confidence maps onto the 3D shape, which are then aggregated through a CRF~\cite{kalogerakis20173d}.
Our approach joins together many of these ideas, leveraging the power of a holistic 3D representation along with features from 2D information by combining them through their explicit spatial mapping.

\section{Method Overview}
Our approach infers 3D object bounding box locations, class labels, and semantic instance masks on a per-voxel basis in an end-to-end fashion.
To this end, we propose a neural network that jointly learns features from both geometry and RGB input.
In the following, we refer to bounding box regression and object classification as object detection, and semantic instance mask segmentation for each object as mask prediction.

In Sec.~\ref{sec:trainingdata}, we first introduce the data representation and  training data that is used by our approach. 
Here, we consider synthetic ground truth data from SUNCG~\cite{song2017ssc}, as well as manually-annotated real-world data from ScanNetV2~\cite{dai2017scannet}.
In Sec.~\ref{sec:architecture}, we present the neural network architecture of our \OURS{} approach.
Our architecture is composed of several parts; on the one hand, we have a series of 3D convolutions that operate in voxel grid space of the scanned 3D data. 
On the other hand, we learn 2D features that we backproject into the voxel grid where we join the features and thus jointly learn from both geometry and RGB data.
These features are used to detect object instances; that is, associated bounding boxes are regressed through a 3D-RPN and class labels are predicted for each object following a 3D-ROI pooling layer.
For each detected object, features from both the 2D color and 3D geometry are forwarded into a per-voxel instance mask network.
Detection and per-voxel instance mask prediction are trained in an end-to-end fashion.
In Sec.~\ref{sec:training}, we describe the training and implementation details of our approach, and in Sec.~\ref{sec:results}, we evaluate our approach.

\begin{figure*}[tp]
\begin{center}
\includegraphics[width=0.95\linewidth]{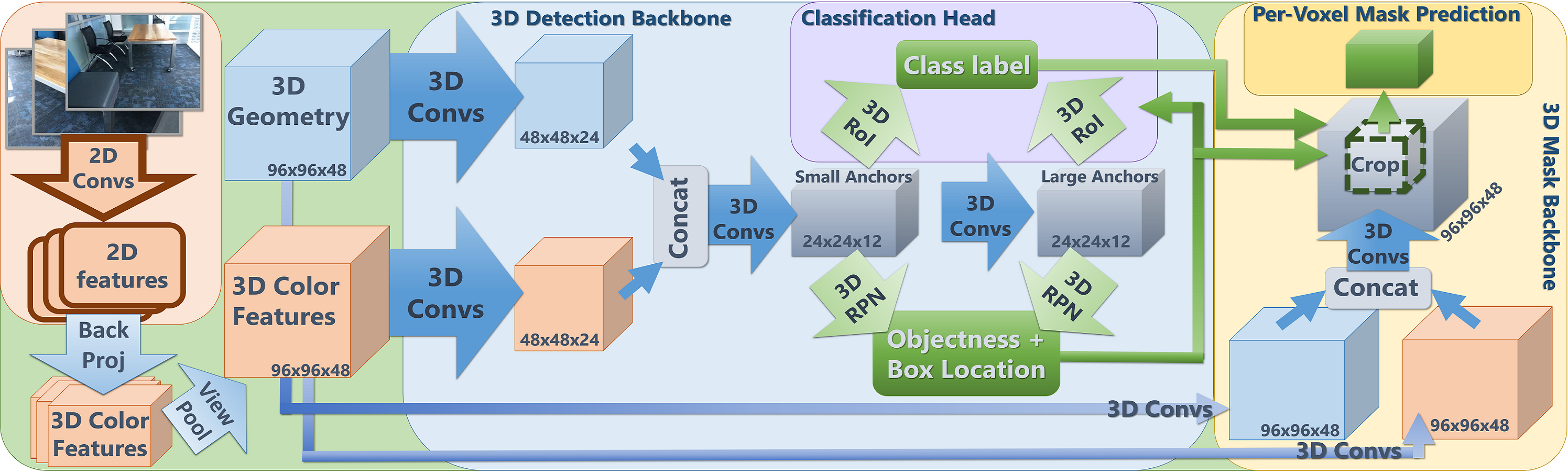}
\end{center}
   \vspace{-0.5cm}
   \caption{\OURS{} network architecture. Our architecture is composed of a 3D detection and a 3D mask pipeline. Both 3D geometry and 2D color images are taken as input and used to jointly learn semantic features for object detection and instance segmentation. From the 3D detection backbone, color and geometry features are used to propose the object bounding boxes and their class labels through a 3D-RPN and a 3D-RoI layer. The mask backbone also uses color and geometry features, in addition to the 3D detection results, to predict per-voxel instance masks inside the 3D bounding box. }
   \vspace{-0.5cm}
\label{fig:network}
\end{figure*}

\section{Training Data}
\label{sec:trainingdata}

\paragraph{Data Representation}
We use a truncated sign distance field (TSDF) representation to encode the reconstructed geometry of the 3D scan inputs.
The TSDF is stored in a regular volumetric grid with truncation of $3$ voxels.
In addition to this 3D geometry, we also input spatially associated RGB images.
This is feasible since we know the mapping between each image pixel with voxels in the 3D scene grid based on the 6 degree-of-freedom (DoF) poses from the respective 3D reconstruction algorithm.

For the training data, we subdivide each 3D scan into chunks of 4.5m $\times$ 4.5m $\times$ 2.25m, and use a resolution of $96\times96\times48$ voxels per chunk (each voxel stores a TSDF value); i.e., our effective voxel size is $\approx4.69$cm$^3$.
In our experiments, for training, we associate 5 RGB images at a resolution of 328x256 pixels in every chunk, with training images selected based on the average voxel-to-pixel coverage of the instances within the region.

Our architecture is fully-convolutional (see Sec.~\ref{sec:architecture}), which allows us to run our method over entire scenes in a single shot for inference.
Here, the xy-voxel resolution is derived from a given test scene's spatial extent.
The z (height) of the voxel grid is fixed to 48 voxels (approximately the height of a room), with the voxel size also fixed at 4.69cm$^3$.
Additionally, at test time, we use all RGB images available for inference.
In order to evaluate our algorithm, we use training, validation, test data from synthetic and real-world RGB-D scanning datasets.

\paragraph{Synthetic Data}
For synthetic training and evaluation, we use the SUNCG~\cite{song2017ssc} dataset.
We follow the public train/val/test split, using 5519 train, 40 validation, and 86 test scenes (test scenes are selected to have total volume $<600$m$^3$).
From the train and validation scenes, we extract $97,918$ train chunks and $625$ validation chunk. 
Each chunk contains an average of $\approx 4.3$ object instances.
At test time, we take the full scan data of the 86 test scenes.

In order to generate partial scan data from these synthetic scenes, we virtually render them, storing both RGB and depth frames.
Trajectories are generated following the virtual scanning approach of ~\cite{dai2018scancomplete}, but adapted to provide denser camera trajectories to better simulate real-world scanning scenarios.
Based on these trajectories, we then generate partial scans as TSDFs through volumetric fusion~\cite{curless1996volumetric}, and define the training data RGB-to-voxel grid image associations based on the camera poses.
We use $23$ class categories for instance segmentation, defined by their NYU40 class labels; these categories are selected for the most frequently-appearing object types, ignoring the wall and floor categories which do not have well-defined instances.

\paragraph{Real-world Data}

For training and evaluating our algorithm on real-world scenes, we use the ScanNetV2~\cite{dai2017scannet} dataset.
This dataset contains RGB-D scans of 1513 scenes, comprising $\approx$2.5 million RGB-D frames.
The scans have been reconstructed using BundleFusion~\cite{dai2017bundlefusion}; both 6 DoF pose alignments and reconstructed models are available. 
Additionally, each scan contains manually-annotated object instance segmentation masks on the 3D mesh.
From this data, we derive 3D bounding boxes which we use as constraints for our 3D region proposal. 

We follow the public train/val/test split originally proposed by ScanNet of 1045 (train), 156 (val), 312 (test) scenes, respectively.
From the train scenes, we extract 108241 chunks, and from the validation scenes, we extract 995 chunks.
Note that due to the smaller number of train scans available in the ScanNet dataset, we augment the train scans to have $4$ rotations each.
We adopt the same $18$-class label set for instance segmentation as proposed by the ScanNet benchmark.

Note that our method is agnostic to the respective dataset as long as semantic RGB-D instance labels are available.

\section{Network Architecture}
\label{sec:architecture}

Our network architecture is shown in Fig.~\ref{fig:network}.
It is composed of two main components, one for detection, and one for per-voxel instance mask prediction; each of these pipelines has its own feature extraction backbone.
Both backbones are composed of a series of 3D convolutions, taking the 3D scan geometry along with the back-projected RGB color features as input.
We detail the RGB feature learning in Sec.~\ref{sec:backproj} and the feature backbones in Sec.~\ref{sec:backbones}.
The learned 3D features of the detection and mask backbones are then fed into the classification and the voxel-instance mask prediction heads, respectively.

The object detection component of the network comprises the detection backbone, a 3D region proposal network (3D-RPN) to predict bounding box locations, and a 3D-region of interest (3D-RoI) pooling layer followed by classification head. 
The detection backbone outputs features which are input to the 3D-RPN and 3D-RoI to predict bounding box locations and object class labels, respectively.
The 3D-RPN is trained by associating predefined anchors with ground-truth object annotations; here, a per-anchor loss defines whether an object exists for a given anchor. If it does, a second loss regresses the 3D object bounding box; if not, no additional loss is considered.
In addition, we classify the the object class of each 3D bounding box.
For the per-voxel instance mask prediction network (see Sec.~\ref{sec:mask}), we use both the input color and geometry as well as the predicted bounding box location and class label.
The cropped feature channels are used to create a mask prediction which has $n$ channels for the $n$ semantic class labels, and the final mask prediction is selected from these channels using the previously predicted class label. 
We optimize for the instance mask prediction using a binary cross entropy loss.
Note that we jointly train the backbones, bounding box regression, classification, and per-voxel mask predictions end-to-end; see Sec.~\ref{sec:training} for more detail.
In the following, we describe the main components of our architecture design, for more detail regarding exact filter sizes, etc., we refer to the supplemental material.

\subsection{Back-projection Layer for RGB Features}
\label{sec:backproj}

In order to jointly learn from RGB and geometric features, one could simply assign a single RGB value to each voxel.
However, in practice, RGB image resolutions are significantly higher than the available 3D voxel resolution due to memory constraints.
This 2D-3D resolution mismatch would make learning from a per-voxel color rather inefficient.
Inspired by the semantic segmentation work of Dai et al.~\cite{dai20183dmv}, we instead leverage a series of 2D convolutions to summarize RGB signal in image space.
We then define a back-projection layer and map these features on top of the associated voxel grid, which are then used for both object detection and  instance segmentation.

To this end, we first pre-train a 2D semantic segmentation network based on the ENet architecture~\cite{paszke2016enet}.
The 2D architecture takes single $256\times328$ RGB images as input, and is trained on a semantic classification loss using the NYUv2 40 label set.
From this pre-trained network, we extract a feature encoding of dimension $32\times 41$ with $128$ channels from the encoder.
Using the corresponding depth image, camera intrinsics, and 6DoF poses, we then  back-project each of these features back to the voxel grid (still 128 channels); the projection is from 2D pixels to 3D voxels.
In order to combine features from multiple views, we perform view pooling through an element-wise max pooling over all RGB images available. 

For training, the voxel volume is fixed to $96\times96\times48$ voxels, resulting in a $128\times96\times96\times48$ back-projected RGB feature grid in 3D; here, we use 5 RGB images for each training chunk (with image selection based on average 3D instance coverage). 
At test time, the voxel grid resolution is dynamic, given by the spatial extent of the environment; here, we use all available RGB images.
The grid of projected features is processed by a set of 3D convolutions and is subsequently merged with the geometric features. 

In ScanNet~\cite{dai2017scannet}, the camera poses and intrinsics are provided; we use them directly for our back-projection layer. 
For SUNCG~\cite{song2017ssc}, extrinsics and intrinsics are given by the virtual scanning path.
Note that our method is agnostic to the used 2D network architecture.

\subsection{3D Feature Backbones}
\label{sec:backbones}

For jointly learning geometric and RGB features for both instance detection and segmentation, we propose two 3D feature learning backbones.
The first backbone generates features for detection, and takes as input the 3D geometry and back-projected 2D features (see Sec.~\ref{sec:backproj}). 

Both the geometric input and RGB features are processed symmetrically with a 3D ResNet block before joining them together through concatenation.
We then apply a 3D convolutional layer to reduce the spatial dimension by a factor of 4, followed by a 3D ResNet block (e.g., for an input train chunk of  $96\times96\times48$, we obtain a features of size $24\times24\times12$).
We then apply another 3D convolutional layer, maintaining the same spatial dimensions, to provide features maps with larger receptive fields.
We define anchors on these two feature maps, splitting the anchors into `small' and `large' anchors (small anchors $<1$m$^3$), with small anchors associated with the first feature map of smaller receptive field and large anchors associated with the second feature map of larger receptive field. 
For selecting anchors, we apply k-means algorithm (k=14) on the ground-truth 3D bounding boxes in first 10k chunks. 
These two levels of features maps are then used for the final steps of object detection: 3D bounding box regression and classification.

The instance segmentation backbone also takes the 3D geometry and the back-projected 2D CNN features as input. 
The geometry and color features are first processed independently with two 3D convolutions, and then concatenated channel-wise and processed with another two 3D convolutions to produce a mask feature map prediction.
Note that for the mask backbone, we maintain the same spatial resolution through all convolutions, which we found to be critical for obtaining high accuracy for the voxel instance predictions.
The mask feature map prediction is used as input to predict the final instance mask segmentation.

In contrast to single backbone, we found that this two-backbone structure both converged more easily and produced significantly better instance segmentation performance (see Sec.~\ref{sec:training} for more details about the training scheme for the backbones).

\subsection{3D Region Proposals and 3D-RoI Pooling for Detection}
\label{sec:detection}

Our 3D region proposal network (3D-RPN) takes input features from the detection backbone to predict and regress 3D object bounding boxes. 
From the detection backbone we obtain two feature maps for small and large anchors, which are separately processed by the 3D-RPN.
For each feature map, the 3D-RPN uses a $1\times1\times1$ convolutional layer  to reduce the channel dimension to $2\times N_\textrm{anchors}$, where $N_\textrm{anchors}=(3, 11)$ for small and large anchors, respectively. 
These represent the positive and negative scores of objectness of each anchor. 
We apply a non-maximum suppression on these region proposals based on their objectness scores.
The 3D-RPN then uses another $1\times1\times1$ convolutional layer to predict feature maps of $6\times N_\textrm{anchors}$, which represent the 3D bounding box locations as $(\Delta_x, \Delta_y, \Delta_z, \Delta_w, \Delta_h, \Delta_l)$, defined in Eq.~\ref{eq:bbox_loss}.

In order to determine the ground truth objectiveness and associated 3D bounding box locations of each anchor during training, we perform anchor association.
Anchors are associated with ground truth bounding boxes by their IoU: if the IoU $>0.35$, we consider an anchor to be positive (and it will be regressed to the associated box), and if the IoU $<0.15$, we consider an anchor to be negative (and it will not be regressed to any box).
We use a two-class cross entropy loss to measure the objectiveness, and for the bounding box regression we use a Huber loss on the prediction $(\Delta_x, \Delta_y, \Delta_z, \Delta_w, \Delta_h, \Delta_l)$  against the log ratios of the ground truth box and anchors $(\Delta_x^{gt}, \Delta_y^{gt}, \Delta_z^{gt}, \Delta_w^{gt}, \Delta_h^{gt}, \Delta_l^{gt})$, where 

\setlength{\abovedisplayskip}{-10pt}
\setlength{\belowdisplayskip}{0pt}
\setlength{\abovedisplayshortskip}{-10pt}
\setlength{\belowdisplayshortskip}{0pt}
\begin{footnotesize}
\begin{equation} \label{eq:bbox_loss}
\Delta_x = \frac{\mu - \mu_{anchor}}{\phi_{anchor}}  \;\;\;\;\;\;
\Delta_w = \ln(\frac{\phi}{\phi_{anchor}})  
\end{equation}
\end{footnotesize}
where $\mu$ is the box center point and $\phi$ is the box width.

Using the predicted bounding box locations, we can then crop out the respective features from the global feature map.
We then unify these cropped features to the same dimensionality using our 3D Region of Interest (3D-RoI) pooling layer.
This 3D-RoI pooling layer pools the cropped feature maps into $4\times4\times4$ blocks through max pooling operations.
These feature blocks are then linearized for input to object classification, which is performed with an MLP.

\vspace{-0.1cm}
\subsection{Per-Voxel 3D Instance Segmentation}
\label{sec:mask}

We perform instance mask segmentation using a separate mask backbone, which similarly as the detection backbone, takes as input the 3D geometry and projected RGB features.
However, for mask prediction, the 3D convolutions maintain the same spatial resolutions, in order to maintain spatial correspondence with the raw inputs, which we found to significantly improve performance.
We then use the predicted bounding box location from the 3D-RPN to crop out the associated mask features from the mask backbone, and compute a final mask prediction with a 3D convolution to reduce the feature dimensionality to $n$ for $n$ semantic class labels; the final mask prediction is the $c^{th}$ channel for predicted object class $c$.
During training, since predictions from the detection pipeline can be wrong, we only train on predictions whose predicted bounding box overlaps with the ground truth bounding box with at least $0.5$ IoU.
The mask targets are defined as the ground-truth mask in the overlapping region of the ground truth box and proposed box.
\setlength{\tabcolsep}{3pt}
\begin{table*}[tp]
    \centering
     \resizebox{\textwidth}{!}{
     \begin{tabular}{l|ccccccccccccccccccccccc||c}\specialrule{1.3pt}{0.0pt}{0.1pt}
     & cab & bed & chair & sofa & tabl & door & wind & bkshf & cntr & desk & shlf & curt & drsr & mirr & tv & nigh & toil & sink & lamp & bath & ostr & ofurn & oprop & \textbf{avg}\\ \hline
     Seg-Cluster &16.8&16.2&15.6&11.8&14.5&10.0&11.7&27.2&20.0&25.7&10.0&0.0&15.0&0.0&20.0&27.8&39.5&22.9&10.7&38.9&10.4&0.0&12.3 & 16.4\\
     Mask R-CNN~\cite{he2017mask} & 14.9&19.0&19.5&13.5&12.2&11.7&14.2&35.0&15.7&18.3&13.7&0.0&24.4&{\bf 23.1}&{\bf 26.0}&{\bf 28.8}&51.2&28.1&14.7&32.2&11.4&10.7&19.5 & 19.9\\
     SGPN~\cite{wang2018sgpn} & 18.6&39.2&28.5&46.5&26.7&{\bf 21.8}&{\bf 15.9}&0.0&24.9&23.9&16.3&{\bf 20.8}&15.1&10.7&0.0&17.7&35.1&37.0&{\bf 22.9}&34.2&17.7&31.5&13.9 & 22.5\\\hline
     Ours(geo only) & 23.2&{\bf 78.6}&47.7&63.3&37.0&19.6&0.0&0.0&21.3&34.4&16.8&0.0&16.7&0.0&10.0&22.8&{\bf 59.7}&49.2&10.0&77.2&10.0&0.0&{\bf 19.3} & 26.8\\ 
     Ours(geo+1view) & 22.2&70.8&48.5&{\bf 66.6}&{\bf 44.4}&10.0&0.0&{\bf 63.9}&25.8&32.2&17.8&0.0&25.3&0.0&0.0&14.7&37.0&55.5&20.5&58.2&{\bf 18.0}&20.0&17.9 & 29.1\\ 
     Ours(geo+3views) & {\bf 26.5}&78.4&48.2&59.5&42.8&26.1&0.0&30.0&22.7&{\bf 39.4}&17.3&0.0&{\bf 36.2}&0.0&10.0&10.0&37.0&50.8&16.8&{\bf 59.3}&10.0&{\bf 36.4}&17.8 & 29.4\\ 
     Ours(geo+5views) & 20.5&69.4&{\bf 56.2}&64.5&43.8&17.8&0.0&30.0&{\bf 32.3}&33.5&{\bf 21.0}&0.0&34.2&0.0&10.0&20.0&56.7&{\bf 56.2}&17.6&56.2&10.0&35.5&17.8 & {\bf 30.6}\\ 
    \specialrule{1.3pt}{0.1pt}{0pt}
    \end{tabular}
    }
     \vspace{-0.3cm}
    \caption{3D instance segmentation on synthetic scans from SUNCG~\cite{song2017ssc}. We evaluate the mean average precision with IoU threshold of 0.25 over 23 classes. Our joint color-geometry feature learning enables us to achieve more accurate instance segmentation performance.
    }
     \vspace{-0.3cm}
    \label{tab:suncg_instance}
\end{table*}

\begin{table*}[bp]
    \centering
     \resizebox{\textwidth}{!}{
     \begin{tabular}{l|cccccccccccccccccc||c}\specialrule{1.3pt}{0.0pt}{0.1pt}
     & cab & bed & chair & sofa & tabl & door & wind & bkshf & pic & cntr & desk & curt & fridg & showr & toil & sink & bath & ofurn & \textbf{avg}\\ \hline
     Mask R-CNN~\cite{he2017mask} &
     5.3&0.2&0.2&10.7&2.0&4.5&0.6&0.0&\textbf{23.8}&0.2&0.0&2.1&6.5&0.0&2.0&1.4&33.3&2.4 & 5.8\\
     SGPN~\cite{wang2018sgpn} & 6.5&39.0&27.5&35.1&16.8&8.7&13.8&16.9&1.4&2.9&0.0&6.9&2.7&0.0&43.8&11.2&20.8&4.3 & 14.3 \\
     MTML & 2.7&\textbf{61.4}&39.0&50.0&10.5&10.0&0.3&\textbf{33.7}&0.0&0.0&0.1&\textbf{11.8}&16.7&14.3&57.0&4.6&66.7&2.8& 21.2 \\
     3D-BEVIS~\cite{Elich19CoRR}&3.5&56.6&39.4&60.4&18.1&9.9&17.1&7.6&2.5&2.7&9.8&3.5&9.8&37.5&85.4&12.6&66.7&3.0& 24.8 \\
    R-PointNet~\cite{yi2018gspn}&\textbf{34.8}&40.5&\textbf{58.9}&39.6&\textbf{27.5}&28.3&\textbf{24.5}&31.1&2.8&\textbf{5.4}&\textbf{12.6}&6.8&21.9&21.4&82.1&\textbf{33.1}&50.0&\textbf{29.0}& 30.6 \\ \hline 
    \OURS{} (Ours) & 13.4&55.4&58.7&\textbf{72.8}&22.4&\textbf{30.7}&18.1&31.9&0.6&0.0&12.1&0.0&\textbf{54.1}&\textbf{100.0}&\textbf{88.9}&4.5&\textbf{66.7}&21.0 & \textbf{36.2} \\ 
    \specialrule{1.3pt}{0.1pt}{0pt}
    \end{tabular}
    }
     \vspace{-0.3cm}
    \caption{Instance segmentation results on the official ScanNetV2 3D semantic instance benchmark (hidden test set). Our final model (geo+5views) significantly outperforms previous (Mask R-CNN, SGPN) and concurrent (MTML, 3D-BEVIS, R-PointNet) state-of-the-art methods in mAP@0.5. ScanNetV2 benchmark data accessed on 12/17/2018. 
    }
     \vspace{-0.4cm}
    \label{tab:scannet_benchmark}
\end{table*}

\section{Training}
\label{sec:training}

To train our model, we first train the detection backbone and 3D-RPN. After pre-training these parts, we add the 3D-RoI pooling layer and object classification head, and train these end-to-end. Then, we add the per-voxel instance mask segmentation network along with the associated backbone. 
In all training steps, we always keep the previous losses (using 1:1 ratio between all losses), and train everything end-to-end.
We found that a sequential training process resulted in more stable convergence and higher accuracy.

We use an SGD optimizer with learning rate 0.001, momentum 0.9 and batch size 64 for 3D-RPN, 16 for classification, 16 for mask prediction. The learning rate is divided by 10 every 100k steps. 
We use a non-maximum suppression for proposed boxes with threshold of 0.7 for training and 0.3 for test.
Our network is implemented with PyTorch and runs on a single Nvidia GTX1080Ti GPU. The object detection components of the network are trained end-to-end for 10 epochs ($\approx 24$ hours). After adding in the mask backbone, we train for an additional 5 epochs ($\approx 16$ hours).
For mask training, we also use ground truth bounding boxes to augment the learning procedure.

\section{Results}
\vspace{-0.1cm}

\label{sec:results}
We evaluate our approach on both 3D detection and instance segmentation predictions, comparing to several state-of-the-art approaches, on synthetic scans of SUNCG~\cite{song2017ssc} data and real-world scans from the ScanNetV2 dataset~\cite{dai2017scannet}.
To compare to previous approaches that operate on single RGB or RGB-D frames (Mask R-CNN~\cite{he2017mask}, Deep Sliding Shapes~\cite{song2015deep}, Frustum PointNet~\cite{qi2017frustum}), we first obtain predictions on each individual frame, and then merge all predictions together in the 3D space of the scene, merging predictions if the predicted class labels match and the IoU $>0.5$.
We further compare to SGPN~\cite{wang2018sgpn} which performs instance segmentation on 3D point clouds.
For both detection and instance segmentation tasks, we project all results into a voxel space of $4.69$cm voxels and evaluate them with a mean average precision metric.
We additionally show several variants of our approach for learning from both color and geometry features, varying the number of color views used during training. 
We consistently find that training on more color views improves both the detection and instance segmentation performance.

\begin{figure*}[h]
\begin{center}
\includegraphics[width=0.92\linewidth]{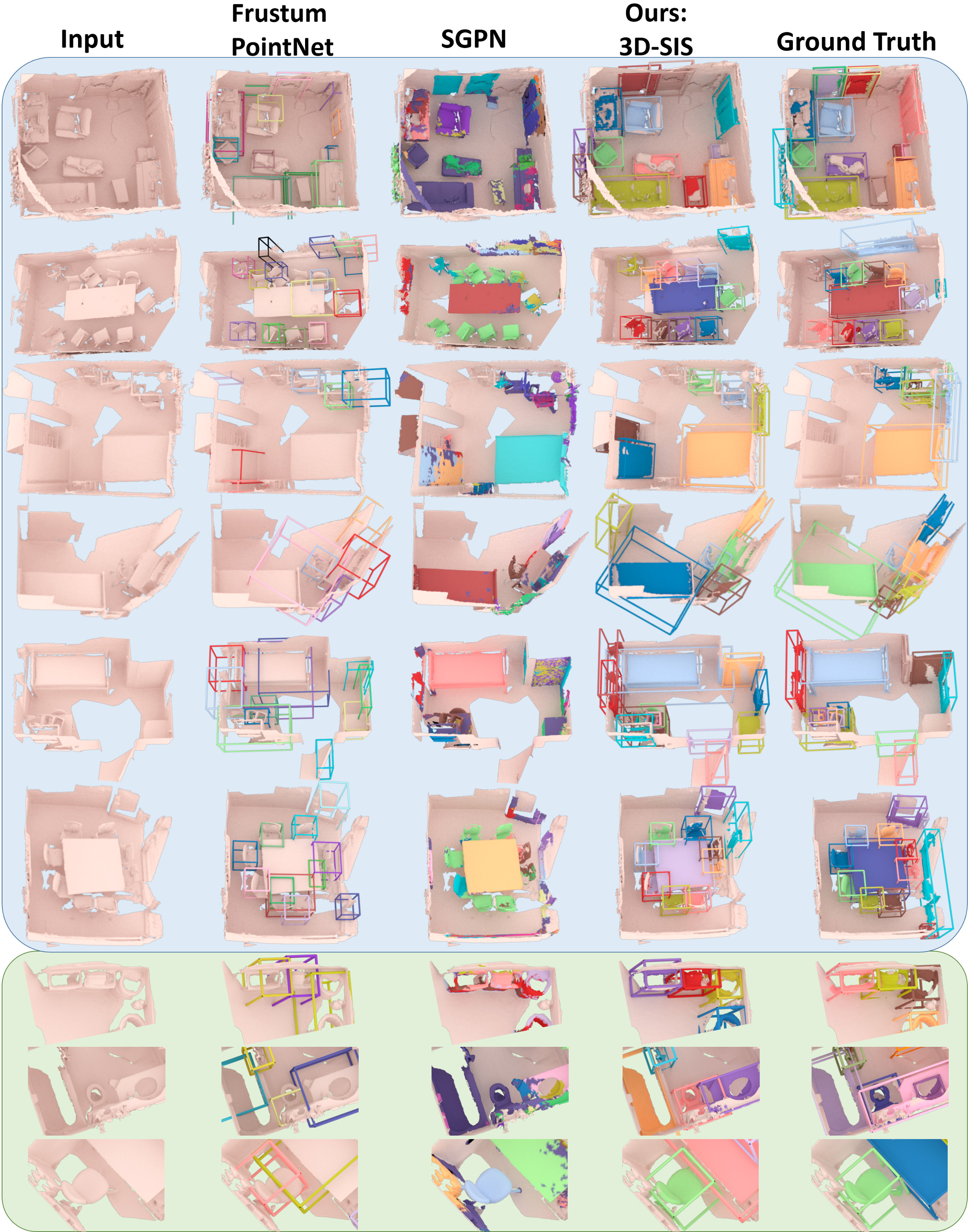}
\end{center}
   \vspace{-0.6cm}
   \caption{Qualitative comparison of 3D object detection and instance segmentation on ScanNetV2~\cite{dai2017scannet} (full scans above; close-ups below). Our joint color-geometry feature learning combined with our fully-convolutional approach to inference on full test scans at once enables more accurate and semantically coherent predictions.
   Note that different colors represent different instances, and the same instances in the ground truth and predictions are not necessarily the same color.}
   \vspace{-0.5cm}
\label{fig:result_comparison}
\end{figure*}

\begin{table*}[h!]
    \centering
     \resizebox{\textwidth}{!}{
     \begin{tabular}{l|cccccccccccccccccc||c}\specialrule{1.3pt}{0.0pt}{0.1pt}
     & cab & bed & chair & sofa & tabl & door & wind & bkshf & pic & cntr & desk & curt & fridg & showr & toil & sink & bath & ofurn & \textbf{avg}\\ \hline
     Seg-Cluster & 11.8&13.5&18.9&14.6&13.8&11.1&11.5&11.7&0.0&13.7&12.2&12.4&11.2&18.0&19.5&18.9&16.4&12.2 & 13.4\\
     Mask R-CNN~\cite{he2017mask} & 15.7&15.4&16.4&16.2&14.9&12.5&11.6&11.8&\textbf{19.5}&13.7&14.4&14.7&21.6&18.5&25.0&24.5&24.5&16.9 & 17.1\\
     SGPN~\cite{wang2018sgpn} & 20.7&31.5&31.6&40.6&\textbf{31.9}&16.6&\textbf{15.3}&13.6&0.0&17.4&14.1&22.2&0.0&0.0&72.9&\textbf{52.4}&0.0&18.6 & 22.2 \\ \hline
    Ours(geo only) & 22.1&48.2&64.4&52.2&16.0&13.4&0.0&17.2&0.0&20.7&17.4&13.9&23.6&33.0&45.2&47.7&61.3&14.6 & 28.3 \\ 
    Ours(geo+1view) & 25.4&60.3&\textbf{66.2}&52.1&31.7&27.6&10.1&16.9&0.0&21.4&30.9&18.4&22.6&16.0&70.5&44.5&37.5&20.0 & 31.8 \\ 
    Ours(geo+3views) & 28.3&52.3&65.0&\textbf{66.5}&31.4&\textbf{27.9}&10.1&\textbf{17.9}&0.0&20.3&\textbf{36.3}&20.1&28.1&31.0&68.6&41&\textbf{66.8}&\textbf{24.0} & 35.3\\ 
    Ours(geo+5views) & \textbf{32.0}&\textbf{66.3}&65.3&56.4&29.4&26.7&10.1&16.9&0.0&\textbf{22.1}&35.1&\textbf{22.6}&\textbf{28.6}&\textbf{37.2}&\textbf{74.9}&39.6&57.6&21.1 & \textbf{35.7} \\ 
    \specialrule{1.3pt}{0.1pt}{0pt}
    \end{tabular}
    }
     \vspace{-0.3cm}
    \caption{3D instance segmentation on ScanNetV2~\cite{dai2017scannet} with mAP@0.25 on 18 classes. 
    Our explicit leveraging of spatial mapping between 3D geometry and color features extracted through 2D CNNs enables significantly improved performance.
    }
     \vspace{-0.4cm}
    \label{tab:scannet_instance}
\end{table*}

\subsection{3D Instance Analysis on Synthetic Scans}
We evaluate 3D detection and instance segmentation on virtual scans taken from the synthetic SUNCG dataset~\cite{song2017ssc}, using 23 class categories. 
Table~\ref{tab:suncg_detection} shows 3D detection performance compared to state-of-the-art approaches which operate on single frames. 
Table~\ref{tab:suncg_instance} shows a quantitative evaluation of our approach, the SGPN for point cloud instance segmentation~\cite{wang2018sgpn}, their proposed Seg-Cluster baseline, and Mask R-CNN~\cite{he2017mask} projected into 3D.
For both tasks, our joint color-geometry approach along with a global view of the 3D scenes at test time enables us to achieve significantly improved detection and segmentation results.

\begin{table}[h!]
\small
\vspace{-0.1cm}
    \centering
    \begin{tabular}{l|c|c}\specialrule{1.1pt}{0pt}{0.1pt}
                         & mAP@0.25 & mAP@0.5 \\ \hline
    Deep Sliding Shapes~\cite{song2014sliding}  &   12.8   &  6.2    \\
    Mask R-CNN 2D-3D~\cite{he2017mask}   &   20.4   &  10.5   \\
    Frustum PointNet~\cite{qi2017frustum}     &   24.9   &  10.8   \\ \hline
     Ours -- \OURS\ (geo only)     &   27.8   &  21.9  \\  
     Ours -- \OURS\ (geo+1view)    &   30.9   &  23.8  \\
     Ours -- \OURS\ (geo+3views)   &   31.3   &  24.2  \\
     Ours -- \OURS\ (geo+5views)   &   \textbf{32.2}   &  \textbf{24.7}  \\\specialrule{1.1pt}{0.1pt}{0pt}
    \end{tabular}
     \vspace{-0.1cm}
    \caption{3D detection in SUNCG~\cite{song2017ssc}, using mAP over 23 classes. Our holistic approach and the combination of color and geometric features result in significantly improved detection results over previous approaches which operate on individual input frames.}
     \vspace{-0.5cm}
    \label{tab:suncg_detection}
\end{table}

\subsection{3D Instance Analysis on Real-World Scans}
We further evaluate our approach on ScanNet dataset~\cite{dai2017scannet}, which contains $1513$ real-world scans. For training and evaluation, we use ScanNetV2 annotated ground truth as well as the proposed 18-class instance benchmark. We show qualitative results in Figure~\ref{fig:result_comparison}. In Table~\ref{tab:scannet_detection}, we quantitatively evaluate our object detection against Deep Sliding Shapes and Frustum PointNet, which operate on RGB-D frame, as well as Mask R-CNN~\cite{he2017mask} projected to 3D. Our fully-convolutional approach enabling inference on full test scenes achieves significantly better detection performance. Table~\ref{tab:scannet_instance} shows our 3D instance segmentation in comparison to SGPN instance segmentation~\cite{wang2018sgpn}, their proposed Seg-Cluster baseline, and Mask R-CNN~\cite{he2017mask} projected into 3D. Our formulation for learning from both color and geometry features brings notable improvement over state of the art.

\begin{table}[h!]
\small
    \centering
    \begin{tabular}{l|c|c}\specialrule{1.1pt}{0pt}{0.1pt}
                         & mAP@0.25 & mAP@0.5 \\ \hline
    Deep Sliding Shapes~\cite{song2014sliding}  &   15.2   &  6.8    \\
    Mask R-CNN 2D-3D~\cite{he2017mask}   &   17.3   &  10.5   \\
    Frustum PointNet~\cite{qi2017frustum} &   19.8   &  10.8   \\ \hline
     Ours -- \OURS\ (geo only)     &   27.6   &  16.0  \\  
     Ours -- \OURS\ (geo+1view)    &   35.1   &  18.7  \\
     Ours -- \OURS\ (geo+3views)   &   36.6   &  19.0  \\
     Ours -- \OURS\ (geo+5views)   &   \textbf{40.2}   &  \textbf{22.5}  \\\specialrule{1.1pt}{0.1pt}{0pt}
    \end{tabular}
     \vspace{-0.1cm}
    \caption{3D detection on ScanNetV2~\cite{dai2017scannet}, using mAP over 18 classes. In contrast to previous approaches operating on individual frames, our approach achieves significantly improved performance.
    }    
     \vspace{-0.55cm}
    \label{tab:scannet_detection}
\end{table}

Finally, we evaluate our model on the ScanNetV2 3D instance segmentation benchmark on the hidden test set; see Table~\ref{tab:scannet_benchmark}.
Our final model (geo+5views) significantly outperforms previous (Mask R-CNN~\cite{he2017mask}, SGPN~\cite{wang2018sgpn}) and concurrent (MTML, 3D-BEVIS~\cite{Elich19CoRR}, R-PointNet~\cite{yi2018gspn}) state-of-the-art methods in mAP@0.5. ScanNetV2 benchmark data was accessed on 12/17/2018.
\vspace{-0.15cm}

\section{Conclusion}
\vspace{-0.1cm}
In this work, we introduce \OURS{}, a new approach for 3D semantic instance segmentation of RGB-D scans, which is trained in an end-to-end fashion to detect object instances and infer a per-voxel 3D semantic instance segmentation.
The core of our method is to jointly learn features from RGB and geometry data using multi-view RGB-D input recorded with commodity RGB-D sensors.
The network is fully-convolutional, and thus can run efficiently in a single shot on large 3D environments.
In comparison to existing state-of-the-art methods that typically operate on single RGB frame, we achieve significantly better 3D detection and instance segmentation results, improving on mAP by over 13. We believe that this is an important insight to a wide range of computer vision applications given that many of them now capture multi-view RGB and depth streams; e.g., autonomous cars, AR/VR applications, etc..
\vspace{-0.1cm}

\section*{Acknowledgments}
\vspace{-0.1cm}
This work was supported by a Google Research Grant, an Nvidia Professor Partnership, a TUM Foundation Fellowship, a TUM-IAS Rudolf M{\"o}{\ss}bauer Fellowship, and the ERC Starting Grant \emph{Scan2CAD (804724)}.

{\small
\bibliographystyle{ieee}
\bibliography{egbib}
}

\clearpage
\begin{appendix}
\setlength{\tabcolsep}{3pt}
\begin{figure*}[tp!]
\begin{center}
\vspace{-0.3cm}
\includegraphics[width=0.95\linewidth]{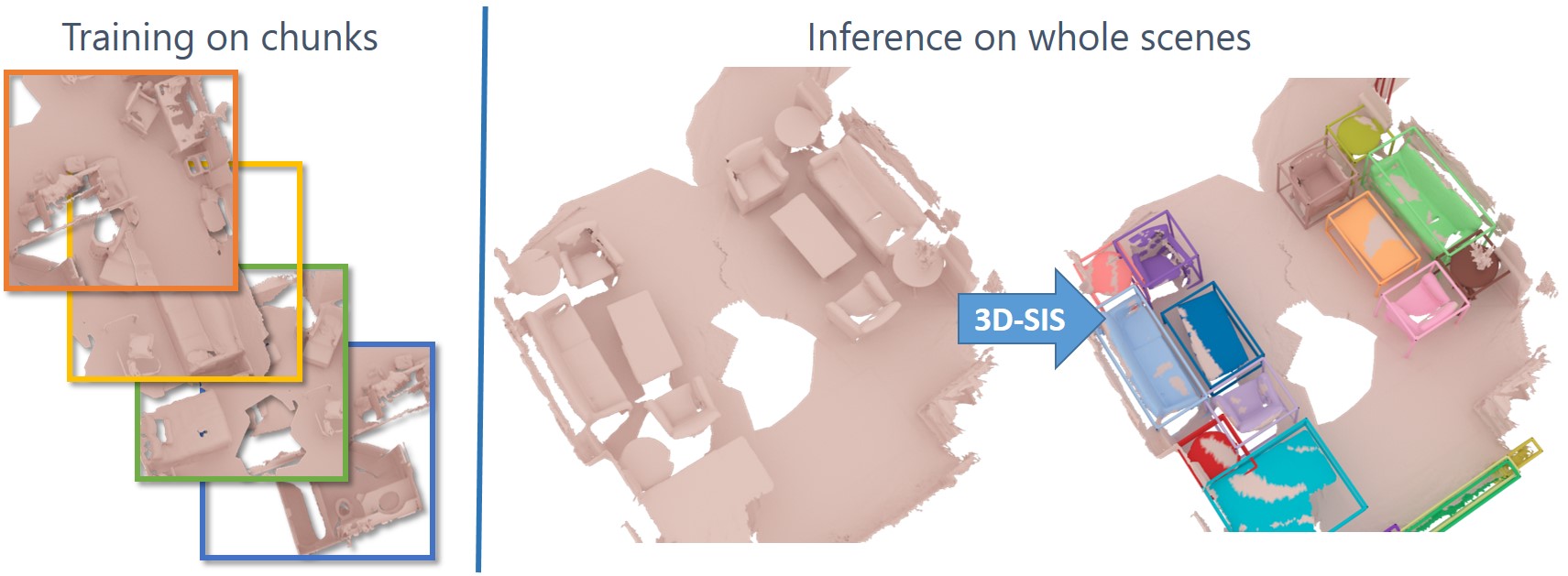}
\end{center}
   \vspace{-0.7cm}
   \caption{\OURS{} trains on chunks of a scene, and leverages fully-convolutional backbone architectures to enable inference on a full scene in a single forward pass, producing more consistent instance segmentation results. }
\label{fig:chunks}
\end{figure*}

In this supplemental document, we describe the details of our \OURS{} network architecture in Section~\ref{sec:architecture_details}.
In Section~\ref{sec:chunk_training}, we describe our training scheme on scene chunks to enable inference on entire test scenes, and finally, in Section~\ref{sec:experiment_details}, we show additional evaluation on the ScanNet~\cite{dai2017scannet} and SUNCG~\cite{song2017ssc} datasets.

\section{Network Architecture}\label{sec:architecture_details}

\begin{table}[hp]
    \centering
     \begin{tabular}{|c|c|} \hline
     small anchors  & big anchors \\ \hline
     (8, 6, 8)     & (12, 12, 40)\\
     (22, 22, 16)  & (8 , 60, 40)\\
     (12, 12, 20)  & (38, 12, 16)\\
                   & (62, 8 , 40)\\
                   & (46, 8 , 20)\\
                   & (46, 44, 20)\\
                   & (14, 38, 16)\\ \hline
    \end{tabular}
    \caption{Anchor sizes (in voxels) used for SUNCG~\cite{song2017ssc} region proposal. Sizes are given in voxel units, with voxel resolution of $\approx 4.69$cm}
    \label{tab:suncg_anchor}
\end{table}
\begin{table}[bp]
    \centering
     \begin{tabular}{|c|c|} \hline
     small anchors  & big anchors \\ \hline
     (8, 8, 9)     & (21, 7,  38)\\
     (14, 14, 11)  & (7,  21, 39)\\
     (14, 14, 20)  & (32, 15, 18)\\
                   & (15, 31, 17)\\
                   & (53, 24, 22)\\
                   & (24, 53, 22)\\
                   & (28, 4,  22)\\
                   & (4,  28, 22)\\
                   & (18, 46, 8) \\ 
                   & (46, 18, 8) \\
                   & (9,  9, 35) \\ \hline
    \end{tabular}
    \caption{Anchor sizes used for region proposal on the ScanNet dataset~\cite{dai2017scannet}. Sizes are given in voxel units, with voxel resolution of $\approx 4.69$cm}
    \label{tab:scannet_anchor}
\end{table}

Table~\ref{tab:network_details} details the layers used in our detection backbone, 3D-RPN, classification head, mask backbone, and mask prediction.
Note that both the detection backbone and mask backbone are fully-convolutional. For the classification head, we use several fully-connected layers; however, due to our 3D RoI-pooling on its input, we can run our entire instance segmentation approach on full scans of varying sizes.

We additionally list the anchors used for the region proposal for our model trained on the ScanNet~\cite{dai2017scannet} and SUNCG~\cite{song2017ssc} datasets in Tables~\ref{tab:scannet_anchor} and \ref{tab:suncg_anchor}, respectively.
Anchors for each dataset are determined through $k$-means clustering of ground truth bounding boxes.
The anchor sizes are given in voxels, where our voxel size is $\approx 4.69$cm.

\begin{table*}[bp]
    \centering
     \begin{tabular}{|c|c|c|c|c|c|c|}\specialrule{1.3pt}{0.0pt}{0.1pt}
     layer name         & input layer   &     type     &    output size                        &   kernel size   &     stride    &   padding  \\ \hline
      geo\_1            & TSDF            & conv3d       & (32, 48, 24, 48)                      & (2, 2, 2)       & (2, 2, 2)     & (0, 0, 0) \\
      geo\_2            & geo\_1             & conv3d       & (32, 48, 24, 48)                      & (1, 1, 1)       & (1, 1, 1)     & (0, 0, 0) \\
      geo\_3            & geo\_2             & conv3d       & (32, 48, 24, 48)                      & (3, 3, 3)       & (1, 1, 1)     & (1, 1, 1) \\
      geo\_4            & geo\_3             & conv3d       & (32, 48, 24, 48)                      & (1, 1, 1)       & (1, 1, 1)     & (0, 0, 0) \\
      geo\_5            & geo\_4             & conv3d       & (32, 48, 24, 48)                      & (1, 1, 1)       & (1, 1, 1)     & (0, 0, 0) \\
      geo\_6            & geo\_5             & conv3d       & (32, 48, 24, 48)                      & (3, 3, 3)       & (1, 1, 1)     & (1, 1, 1) \\
      geo\_7            & geo\_6             & conv3d       & (32, 48, 24, 48)                      & (1, 1, 1)       & (1, 1, 1)     & (0, 0, 0) \\
      geo\_8            & geo\_7            & conv3d       & (64, 24, 12, 24)                      & (2, 2, 2)       & (2, 2, 2)     & (0, 0, 0) \\
      geo\_9            & geo\_1             & conv3d       & (32, 24, 12, 24)                      & (1, 1, 1)       & (1, 1, 1)     & (0, 0, 0) \\
      geo\_10            & geo\_2             & conv3d       & (32, 24, 12, 24)                      & (3, 3, 3)       & (1, 1, 1)     & (1, 1, 1) \\
      geo\_11            & geo\_3             & conv3d       & (64, 24, 12, 24)                      & (1, 1, 1)       & (1, 1, 1)     & (0, 0, 0) \\
      geo\_12            & geo\_4             & conv3d       & (32, 24, 12, 24)                      & (1, 1, 1)       & (1, 1, 1)     & (0, 0, 0) \\
      geo\_13            & geo\_5             & conv3d       & (32, 24, 12, 24)                      & (3, 3, 3)       & (1, 1, 1)     & (1, 1, 1) \\
      geo\_14            & geo\_6             & conv3d       & (64, 24, 12, 24)                      & (1, 1, 1)       & (1, 1, 1)     & (0, 0, 0) \\
      color\_1          & projected 2D features             & conv3d       & (64, 48, 24, 48)                      & (2, 2, 2)       & (2, 2, 2)     & (0, 0, 0) \\
      color\_2          & color\_1              & conv3d       & (32, 48, 24, 48)                      & (1, 1, 1)       & (1, 1, 1)     & (0, 0, 0) \\
      color\_3          & color\_2              & conv3d       & (32, 48, 24, 48)                      & (3, 3, 3)       & (1, 1, 1)     & (1, 1, 1) \\
      color\_4          & color\_3              & conv3d       & (64, 48, 24, 48)                      & (1, 1, 1)       & (1, 1, 1)     & (0, 0, 0) \\
      color\_5          & color\_4              & maxpool3d    & (64, 48, 24, 48)                      & (3, 3, 3)       & (1, 1, 1)     & (1, 1, 1) \\
      color\_6          & color\_5             & conv3d       & (64, 24, 12, 24)                      & (2, 2, 2)       & (2, 2, 2)     & (0, 0, 0) \\
      color\_7          & color\_6              & conv3d       & (32, 24, 12, 24)                      & (1, 1, 1)       & (1, 1, 1)     & (0, 0, 0) \\
      color\_8          & color\_7              & conv3d       & (32, 24, 12, 24)                      & (3, 3, 3)       & (1, 1, 1)     & (1, 1, 1) \\
      color\_9          & color\_8              & conv3d       & (64, 24, 12, 24)                      & (1, 1, 1)       & (1, 1, 1)     & (0, 0, 0) \\
      color\_10          & color\_9              & maxpool3d       & (64, 24, 12, 24)                      & (3, 3, 3)       & (1, 1, 1)     & (1, 1, 1) \\
      concat\_1         & (geo\_14, color\_10)              & concat       & (128, 24, 12, 24)                     &  None           & None          & None      \\
      combine\_1        & concat\_1             & conv3d       & (128, 24, 12, 24)                     & (3, 3, 3)       & (1, 1, 1)     & (1, 1, 1) \\
      combine\_2        & combine\_1              & conv3d       & (64, 24, 12, 24)                      & (1, 1, 1)       & (1, 1, 1)     & (0, 0, 0) \\
      combine\_3        & combine\_2              & conv3d       & (64, 24, 12, 24)                      & (3, 3, 3)       & (1, 1, 1)     & (1, 1, 1) \\
      combine\_4        & combine\_3              & conv3d       & (128, 24, 12, 24)                     & (1, 1, 1)       & (1, 1, 1)     & (0, 0, 0) \\
      combine\_5        & combine\_4              & conv3d       & (64, 24, 12, 24)                      & (1, 1, 1)       & (1, 1, 1)     & (0, 0, 0) \\
      combine\_6        & combine\_5              & conv3d       & (64, 24, 12, 24)                      & (3, 3, 3)       & (1, 1, 1)     & (1, 1, 1) \\
      combine\_7        & combine\_6              & conv3d       & (128, 24, 12, 24)                     & (1, 1, 1)       & (1, 1, 1)     & (0, 0, 0) \\
      combine\_8        & combine\_7              & maxpool3d    & (128, 24, 12, 24)                      & (3, 3, 3)       & (1, 1, 1)     & (1, 1, 1) \\
      rpn\_1            & combine\_7              & conv3d       & (256, 24, 12, 24)                     & (3, 3, 3)       & (1, 1, 1)     & (1, 1, 1) \\
      rpn\_cls\_1       & rpn\_1              & conv3d       & (6, 24, 12, 24)                       & (1, 1, 1)       & (1, 1, 1)     & (0, 0, 0) \\
      rpn\_bbox\_1      & rpn\_1              & conv3d       & (18, 24, 12, 24)                      & (1, 1, 1)       & (1, 1, 1)     & (0, 0, 0) \\
      rpn\_2            & combine\_5              & conv3d       & (256, 24, 12, 24)                     & (3, 3, 3)       & (1, 1, 1)     & (1, 1, 1) \\
      rpn\_cls\_2       & rpn\_2              & conv3d       & (22, 24, 12, 24)                      & (1, 1, 1)       & (1, 1, 1)     & (0, 0, 0) \\
      rpn\_bbox\_2      & rpn\_2              & conv3d       & (66, 24, 12, 24)                      & (1, 1, 1)       & (1, 1, 1)     & (0, 0, 0) \\
      cls\_1            & combine\_7              & FC           & 128x4x4x4 $\rightarrow$ 256           & None            & None          & None      \\
      cls\_2            & cls\_1              & FC           & 256 $\rightarrow$ 256                 & None            & None          & None      \\ 
      cls\_3            & cls\_2              & FC           & 256 $\rightarrow$ 128                 & None            & None          & None      \\ 
      cls\_cls          & cls\_3              & FC           & 128 $\rightarrow$ $N_{cls}$           & None            & None          & None      \\ 
      cls\_bbox         & cls\_3              & FC           & 128 $\rightarrow$  $N_{cls}\times6$   & None            & None          & None      \\ 
      mask\_1    & TSDF           & conv3d       & (64, 96, 48, 96)                       & (3, 3, 3)       & (1,1,1)       & (1,1,1)      \\
      mask\_2    & mask\_1        & conv3d       & (64, 96, 48, 96)                       & (3, 3, 3)       & (1,1,1)       & (1,1,1)      \\
      mask\_3    & mask\_2        & conv3d       & (64, 96, 48, 96)                       & (3, 3, 3)       & (1,1,1)       & (1,1,1)      \\
      mask\_4    & mask\_3        & conv3d       & (64, 96, 48, 96)                       & (3, 3, 3)       & (1,1,1)       & (1,1,1)      \\
      mask\_5    & mask\_4        & conv3d       & (64, 96, 48, 96)                       & (3, 3, 3)       & (1,1,1)       & (1,1,1)      \\
      mask\_6    & mask\_5        & conv3d       & ($N_{cls}$, 96, 48, 96)                & (1, 1, 1)       & (1,1,1)       & (0,0,0)      \\
    \specialrule{1.3pt}{0.1pt}{0pt}
    \end{tabular}
    \caption{\OURS{} network architecture layer specifications.}
    \label{tab:network_details}
\end{table*}

\section{Training and Inference}\label{sec:chunk_training}
In order to leverage as much context as possible from a input RGB-D scan, we leverage fully-convolutional detection and mask backbones to infer instance segmentation on varying-sized scans.
To accommodate memory and efficiency constraints during training, we train on chunks of scans, i.e. cropped volumes out of the scans, which we use to generalize to the full scene at test time (see Figure~\ref{fig:chunks}).
This also enables us to avoid inconsistencies which can arise with individual frame input, with differing views of the same object; with the full view of a test scene, we can more easily predict consistent object boundaries.

\begin{figure*}[tp]
    \centering
    \includegraphics[width=0.90\linewidth,trim={0cm 0.0cm 0 0},clip]{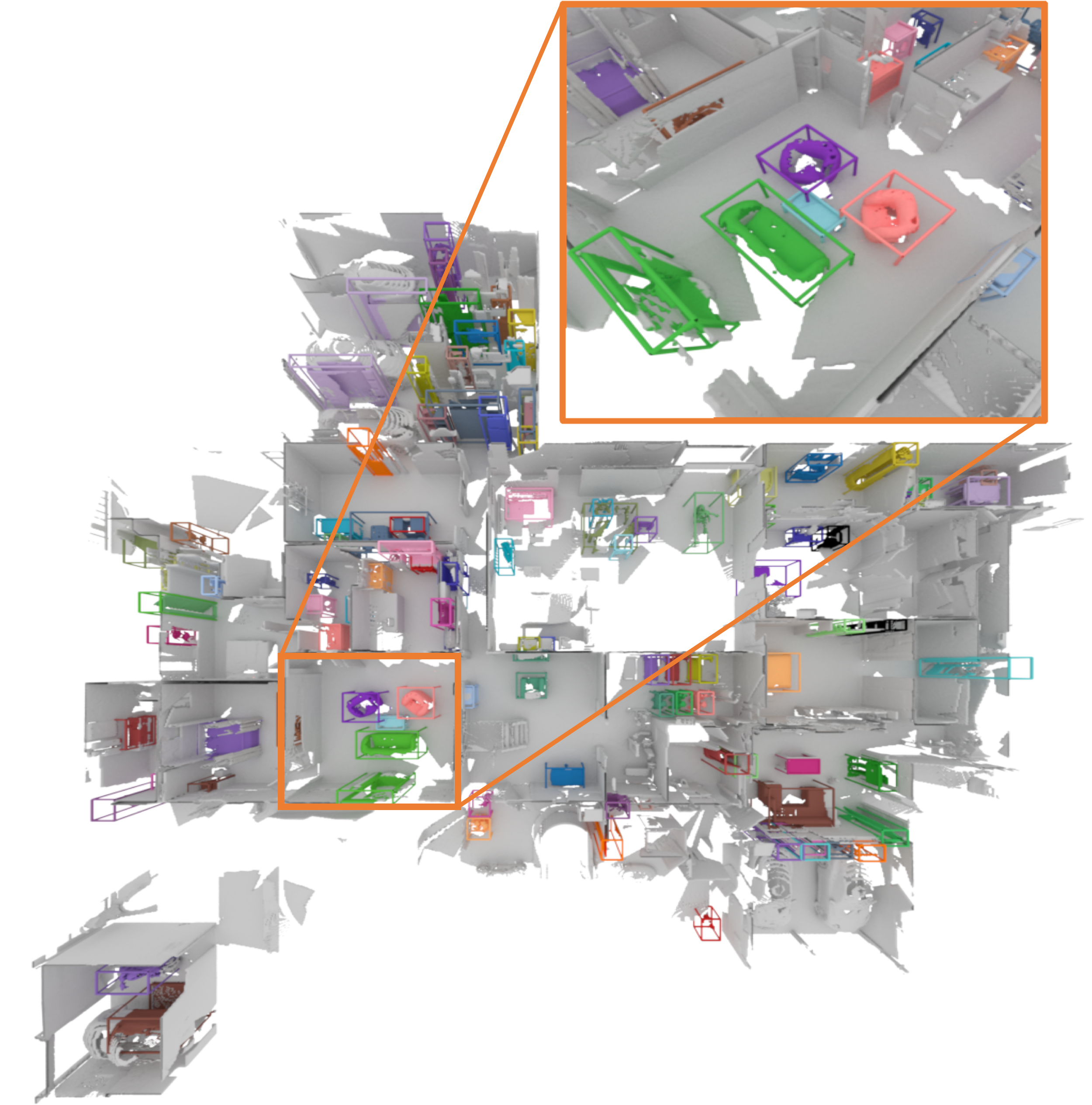}
    \caption{Our fully-convolutional architectures allows testing on a large SUNCG scene (45m x 45m) in about 1 second runtime.}
    \label{fig:suncg_large}
\end{figure*}

The fully-convolutional nature of our methods allows testing on very large scans such as entire floors or buildings in a single forward pass; e.g., most SUNCG scenes are actually fairy large; see Figure~\ref{fig:suncg_large}.

\section{Additional Experiment Details}\label{sec:experiment_details}
We additionally evaluate mean average precision on SUNCG~\cite{song2017ssc} and ScanNetV2~\cite{dai2017scannet} using an IoU threshold of 0.5 in Tables~\ref{tab:suncg_instance_05} and \ref{tab:scannet_instance_05}.
Consistent with evaluation at an IoU threshold of 0.25, our approach leveraging joint color-geometry feature learning and inference on full scans enables significantly better instance segmentation performance. 
We also submit our model the ScanNet Benchmark, and we achieve the state-of-the-art in all three metrics.

\begin{table}[bp!]
    \centering
    \footnotesize
     \begin{tabular}{|l|c|c|}                                                        \hline
                                 & mAP@0.5 & mAP@0.25 \\ \hline
       3D-SIS (only color-1view) & 9.4 & 30.5 \\\hline
       3D-SIS (only color-3view) & 16.5& 35.0 \\\hline
       3D-SIS (only color-5view) & 17.4 & 35.7 \\ \hline
       3D-SIS (only geometry)    & 16.0 & 27.6 \\ \hline
       3D-SIS (one anchor layer) & 12.2 & 33.4\\ \hline \hline
       3D-SIS (final) & \textbf{22.5} & \textbf{40.2} \\ \hline
    \end{tabular}
    \caption{Additional ablation study on ScanNetV2; combination of geometry and color signal complement each other, thus achieving the best performance.}
    \vspace{-0.3cm}
    \label{tab:scannet_instance}
\end{table}

\begin{table*}[tp]
    \centering
     \resizebox{\textwidth}{!}{
     \begin{tabular}{l|cccccccccccccccccc||c}\specialrule{1.3pt}{0.0pt}{0.1pt}
     & cab & bed & chair & sofa & tabl & door & wind & bkshf & pic & cntr & desk & curt & fridg & showr & toil & sink & bath & ofurn & \textbf{avg}\\ \hline
     Seg-Cluster&10.4&11.9&15.5&12.8&12.4&10.1&10.1&10.3&0.0&11.7&10.4&11.4&0.0&13.9&17.2&11.5&14.2&10.5&10.8\\
     Mask R-CNN~\cite{he2017mask}&11.2&10.6&10.6&11.4&10.8&10.3&0.0&0.0&\bf{11.1}&\bf{10.1}&0.0&10.0&12.8&0.0&18.9&13.1&11.8&11.6&9.1\\
     SGPN~\cite{wang2018sgpn}&10.1&16.4&20.2&20.7&14.7&11.1&\bf{11.1}&0.0&0.0&10.0&10.3&12.8&0.0&0.0&48.7&16.5&0.0&0.0&11.3\\ \hline
    Ours(geo only)&11.5&17.5&18.0&26.3&0.0&10.1&0.0&10.3&0.0&0.0&0.0&0.0&\bf{24.4}&21.5&25.0&\bf{17.2}&34.9&10.1&12.6\\ 
    Ours(geo+1view)&12.5&15.0&17.8&23.7&0.0&\bf{19.0}&0.0&11.0&0.0&0.0&10.5&11.1&13.0&19.4&22.5&14.0&\bf{40.5}&10.1&13.3\\ 
    Ours(geo+3views)&14.4&19.9&\bf{48.4}&\bf{37.3}&\bf{16.9}&18.3&0.0&11.0&0.0&0.0&10.5&\bf{13.1}&16.3&15.3&\bf{51.3}&13.0&12.9&\bf{13.4}&17.3\\ 
    Ours(geo+5views)&\bf{19.7}&\bf{37.7}&40.5&31.9&15.9&18.1&0.0&\bf{11.0}&0.0&0.0&\bf{10.5}&11.1&18.5&\bf{24.0}&45.8&15.8&23.5&12.9&\bf{18.7}\\ 
    \specialrule{1.3pt}{0.1pt}{0pt}
    \end{tabular}
    }
     \vspace{-0.3cm}
    \caption{3D instance segmentation on real-world scans from ScanNetV2~\cite{dai2017scannet}.  We evaluate the mean average precision with IoU threshold of 0.5 over 18 classes. 
    Our explicit leveraging of the spatial mapping between the 3D geometry and color features extracted through 2D convolutions enables significantly improved instance segmentation performance.
    }
    \label{tab:scannet_instance_05}
\end{table*}

\begin{table*}[tp!]
    \centering
     \resizebox{\textwidth}{!}{
     \begin{tabular}{l|ccccccccccccccccccccccc||c}\specialrule{1.3pt}{0.0pt}{0.1pt}
     & cab & bed & chair & sofa & tabl & door & wind & bkshf & cntr & desk & shlf & curt & drsr & mirr & tv & nigh & toil & sink & lamp & bath & ostr & ofurn & oprop & \textbf{avg}\\ \hline
     Seg-Cluster &10.1&10.9&10.4&10.1&10.3&0.0&0.0&12.9&10.7&15.2&0.0&0.0&10.0&0&0.0&11.2&26.1&12.1&0&16.5&0&0&10&7.7\\
     Mask R-CNN~\cite{he2017mask} &0.0&10.7&0.0&0.0&0.0&0.0&0.0&0.0&0.0&0.0&0.0&0.0&0.0&\bf{10.8}&\bf{11.4}&10.8&18.8&13.5&0.0&11.5&0.0&0.0&10.7&4.3\\
     SGPN~\cite{wang2018sgpn}&15.3&28.7&23.7&29.7&17.6&15.1&\bf{15.4}&0.0&10.8&16.0&0.0&\bf{10.9}&0.0&0.0&0.0&12.3&33.7&25.9&\bf{19.2}&31.7&0.0&10.4&10.5&14.2 \\ \hline
     Ours(geo only) &12.6&\bf{60.5}&38.6&45.8&21.8&\bf{16.8}&0.0&0.0&10.0&18.5&10.0&0.0&14.0&0.0&0.0&\bf{14.9}&\bf{64.2}&30.8&17.6&35.2&10.0&0.0&\bf{16.9}&19.1\\ 
     Ours(geo+1view)&13.9&42.4&35.3&\bf{52.9}&22&10&0.0&\bf{35.0}&\bf{13.4}&\bf{21.4}&10.0&0.0&13.5&0.0&0.0&10.0&33.8&29.2&\bf{17.7}&\bf{48.3}&10.0&16.9&10.0&19.4\\ 
     Ours(geo+3views)&15.4&58.5&35.5&34.5&\bf{24.4}&16.6&0.0&20.0&10.0&17.6&10.0&0.0&\bf{24.3}&0.0&10.0&10.0&34.6&28.5&15.6&40.7&10.0&\bf{24.9}&15.5&19.8\\ 
     Ours(geo+5views) &\bf{15.5}&43.6&\bf{43.9}&48.1&20.4&10.0&0.0&30.0&10.0&17.4&\bf{10.0}&0.0&14.5&0.0&10.0&10.0&53.5&\bf{35.1}&17.2&39.7&\bf{10.0}&18.9&16.2&\bf{20.6}\\ 
    \specialrule{1.3pt}{0.1pt}{0pt}
    \end{tabular}
    }
     \vspace{-0.3cm}
    \caption{3D instance segmentation on synthetic scans from SUNCG~\cite{song2017ssc}. We evaluate the mean average precision with IoU threshold of 0.5 over 23 classes. Our joint color-geometry feature learning enables us to achieve more accurate instance segmentation performance.
    }
     \vspace{-0.3cm}
    \label{tab:suncg_instance_05}
\end{table*}

We run an additional ablation study to evaluate the impact of the RGB input and the two-level anchor design; see Table.~\ref{tab:scannet_instance}.


\end{appendix}

\end{document}